\documentclass{article}
\usepackage{main_style}

\usepackage{times}
\usepackage{soul}
\usepackage{url}
\usepackage[hidelinks]{hyperref}
\usepackage[utf8]{inputenc}
\usepackage[small]{caption}
\usepackage{graphicx}
\usepackage{amsmath}
\usepackage{booktabs}
\usepackage{shortbold}
\usepackage{xcolor}
\usepackage{placeins}
\usepackage[ruled]{algorithm2e}
\usepackage{verbatim}
\usepackage{lipsum}



\title{Distilling with Performance Enhanced Students}

\author{
Jack Turner$^{*1}$, Elliot J. Crowley$^{*1}$, Valentin Radu$^1$, Jos\'e Cano$^2$, Amos Storkey$^1$, Michael O'Boyle$^1$
\\
$^1$School of Informatics, University of Edinburgh\\
$^2$School of Computing Science, University of Glasgow\\
$^*$Equal Contribution\\
}

\begin{document}

\maketitle

\begin{abstract}
The task of accelerating large neural networks on general purpose hardware has, in recent years, prompted the use of channel pruning to reduce network size. However, the efficacy of pruning based approaches has since been called into question. In this paper, we turn to distillation for model compression---specifically, attention transfer---and develop a simple method for discovering performance enhanced student networks. We combine channel saliency metrics with empirical observations of runtime performance to design more accurate networks for a given latency budget. We apply our methodology to residual and densely-connected networks, and show that we are able to find resource-efficient student networks on different hardware platforms while maintaining very high accuracy. These performance-enhanced student networks achieve up to 10\% boosts in top-1 ImageNet accuracy over their channel-pruned counterparts for the same inference time.
\end{abstract}

\section{Introduction}
\label{intro}

Patience is often said to be a virtue. However, for a deep neural network performing inferences on an edge device one cannot afford to be patient e.g.\ for pedestrian detection. So how do we make our networks faster? A common approach is to use a neural compression technique such as pruning or distillation to reduce the number of parameters a network uses.

In pruning, weights connecting neurons~\cite{lecun1990optimal,han2015learning}, or whole channels~\cite{molchanov2016pruning,theis2018faster,louizos2017bayesian} are removed while a network is fine-tuned. It is preferable to remove channels, as the sparsity introduced by weight pruning is difficult to exploit on general purpose hardware~\cite{sze2017efficient,turner2018characterising}.

\begin{figure}
    \centering
    \includegraphics[width=\linewidth]{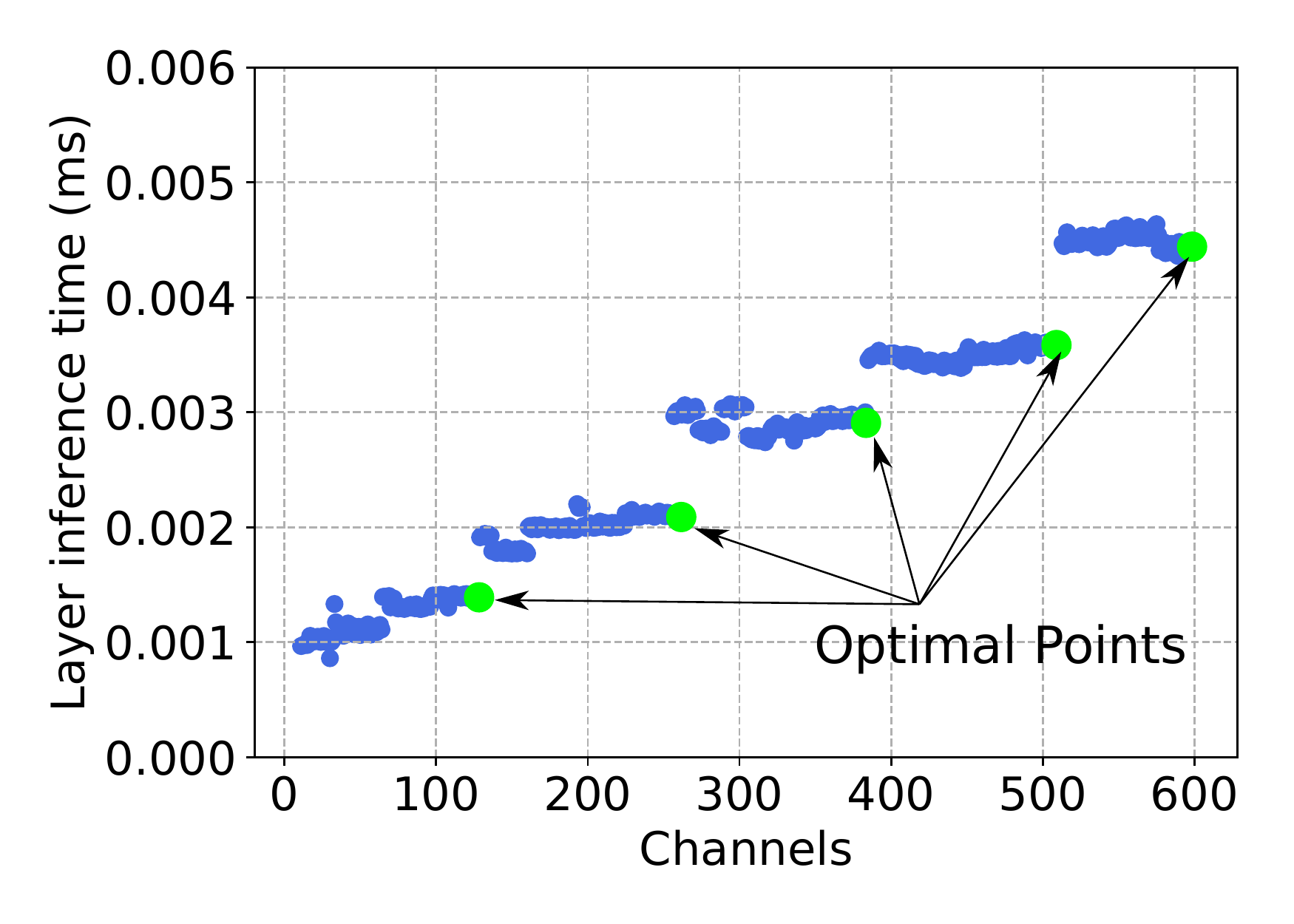}
    \vspace{-0.5cm}
    \caption{Inference time for a layer of a ResNet-34 vs. the number of channels in that layer on an ARM Cortex-A57. Notice the staircase pattern that emerges. For a given inference time it is preferable to pick the points in green to maximize the network's capacity.}
    \label{fig:layer9-gpu}
\end{figure}

\begin{figure*}[t]
    \includegraphics[width=\linewidth]{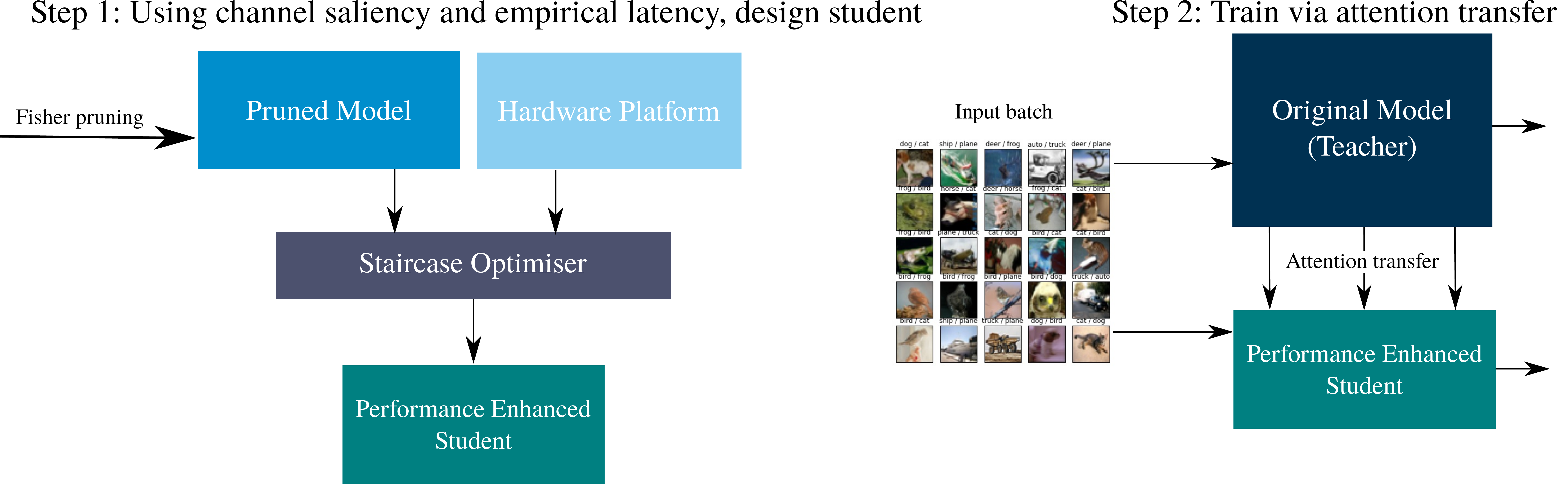}
    \caption{Our student-discovery and optimisation pipeline. In step 1, we use the Fisher-pruned model and our staircase optimiser to design a student network that fits a specific deployment platform. In step 2 we use the original model to train the performance enhanced student using attention transfer.}
    \label{fig:pipeline}
\end{figure*}

However, the efficacy of channel pruning has recently been called into question~\cite{liu2018rethinking,crowley2018pruning}: the common pattern of pruning and fine-tuning is not as effective as simply training the pruned architecture from scratch. Moreover, channel pruning relies on the assumption that reducing the number of activation channels in a layer---and therefore the number of floating point operations (FLOPs) it uses---will linearly reduce inference time, which is often not the case~\cite{yu2017scalpel,netadapt}. In Figure~\ref{fig:layer9-gpu} we show the effect of pruning channels in an early convolutional layer of a ResNet-34~\cite{he2016deep} on the ARM Cortex-A57 CPU. Instead of linear speedup, we observe a~{\bf staircase pattern}, with steps at points that fully utilise the available device parallelism. It is important to note that many of these steps are not related to the typical powers of two that can be leveraged by Single Instruction Multiple Data (SIMD) or Multiple Instruction Multiple Data (MIMD) processors; there are nuances in the space of matrix representations and choice of arithmetic primitives that determine the locations. By ignoring this and using FLOPs as a direct proxy for inference time, pruning techniques therefore often converge on inefficient network architectures.

In distillation, one produces a smaller student network and uses the outputs~\cite{ba2014do,hinton2015distilling} or activations~\cite{romero2014fitnets,zagoruyko2016paying} of the original large network---referred to as the teacher---to aid in its training. The difficulty of this process is picking the right student; it could have reduced width, depth, or even substituted convolutions~\cite{moonshine} but these may not run quickly when deployed on a particular device.

In this work, we propose a novel {\it latency-aware} technique that combines pruning and distillation, combats their flaws, and produces a {\it performance enhanced} student that runs quickly on a target device.

Starting with a large, trained network, we first perform pruning---specifically, Fisher pruning~\cite{theis2018faster}---and then use a latency-aware optimiser to adapt the profile of the pruned network for deployment on a particular device. Finally, we treat this adapted network as a student and distill it with the original large network as a teacher using attention transfer~\cite{zagoruyko2016paying} to recover its performance. This technique is illustrated in Figure~\ref{fig:pipeline}.

By combining channel saliency metrics with empirical observations of hardware performance we develop a new methodology for reshaping neural networks to maximise accuracy for a given latency budget. The existence of stepped inference speedups allows us to increase the capacity of pruned network architectures without affecting total latency, resulting in reduced error and greater resource efficiency.

In Section~\ref{prelim} pruning and distillation are briefly reviewed. We then describe how {\it performance enhanced} students are discovered for a target device (Section~\ref{distill}). In Section~\ref{cifar} our technique is applied to WideResNets~\cite{zagoruyko2016wide} and DenseNets~\cite{huang2017densely} on the CIFAR-10~\cite{cifar} dataset. We show this extends to ResNets trained on ImageNet~\cite{ILSVRC15} in Section~\ref{imagenet}.

The contributions of this paper are as follows:
\begin{itemize}
\item A simple method for identifying performance-optimal layer widths for a given neural network and hardware platform.
\item A new technique for finding performance enhanced architecture reductions for deploying neural networks in resource-constrained scenarios.
\item The study and benchmarking of our methodology against common pruning techniques on typical examples of embedded hardware platforms. 
\end{itemize}


\section{Related Work}

The parameter redundancy in modern neural networks has been well studied~\cite{denil2013predicting}. This is often exploited for inference acceleration in various forms, most commonly through network pruning. This typically comes in two varieties: weight pruning~\cite{han2015learning}  and channel pruning~\cite{he2017channel,molchanov2016pruning,netadapt,he2018amc,theis2018faster}. 

In weight pruning, individual weights (connections between neurons) are zeroed out---usually in stages---while the network is fine-tuned, leaving a sparse set of weight tensors; at present, these are difficult to deploy on general-purpose hardware~\cite{turner2018characterising}. Channel pruning focuses on the removal of neuron structures (or filters) while fine-tuning, leaving the weight tensor smaller but still dense, which requires fewer operations on dense matrices.

The detection of unimportant weights and channels relies on a saliency estimation metric. Magnitude-based metrics e.g.~\cite{li2016pruning} assume that the importance of neurons (or neuronal groups) is implied by their relative absolute norm. However, the assumption that magnitude is correlated to importance has been questioned~\cite{ye2018rethinking}. Another approach is to use a second-order Taylor expansion of the network's loss function to estimate the increase in loss that the removal of each neuronal group would induce~\cite{molchanov2016pruning,theis2018faster}; a technique known as Fisher pruning.

There are however, two major disadvantages with the pruning paradigm. First, pruning often leads to mediocre inference speedups as the resultant architectures do not make efficient use of available hardware characteristics. In some cases, this may directly contradict optimisations that happen lower down in the stack (e.g. pruned channels may simply be padded up to allow for layer fusion). Second, the pruning-and-tuning framework commonly used has recently been shown to be inferior to training the network from scratch~\cite{crowley2018pruning,liu2018rethinking}. Other works have made use of empirical observations of latency to guide their architecture search~\cite{netadapt,he2018amc}, but suffer accuracy costs from reliance on the pruning-and-tuning pattern. 

An alternative means of making a network smaller is through a distillation process~\cite{ba2014do,hinton2015distilling} whereby a small \textit{student} network is trained both on the data and on the outputs of a larger \textit{teacher} network. This has been shown empirically to yield higher accuracy than training on just the data. Alternatively, the student can utilise the teacher's activations~\cite{romero2014fitnets,zagoruyko2016paying} for a greater boost.
Typically the student network is obtained by reducing the channels per layer or by reducing the number of layers of a teacher network, or even replacing the convolutional layers with more efficient convolutional structures~\cite{moonshine}.  However, the structure of these students may be unsuitable for fast inference on specialised devices.


\section{Preliminaries}
\label{prelim}
Here, we briefly summarise the pruning and distillation method used in our acceleration technique.

\subsection{Fisher pruning}

Fisher pruning~\cite{theis2018faster,molchanov2016pruning} is a principled channel pruning technique, whereby the saliency metric is an approximation of the change in error that would occur on the removal of a channel. More formally, consider a single channel of an activation in a network due to some input minibatch of $N$ examples. Let us denote this as $C$: it is an $N \times W \times H$ tensor where $W$ and $H$ are the channel's spatial width and height. Let us refer to the entry corresponding to example $n$ in the mini-batch at location $(i,j)$ as $C_{nij}$. If the network has a loss function $\mathcal{L}$, then we can back-propagate to get the gradient of the loss with respect to this activation channel $\frac{\partial \mathcal{L}}{\partial {C}}$. Let us denote this gradient as $g$ and index it as $g_{nij}$. $\Delta_{c}$ can then be approximated as:

\begin{equation}
\Delta_{c} =\frac{1}{2N} \sum_{n}^{N}\left(- \sum_{i}^{W}\sum_{j}^{H}C_{nij} g_{nij}\right)^2
\end{equation}

This quantity is accumulated as the network is fine-tuned, and then the channel with the lowest $\Delta_{c}$ value is pruned. This pruning-and-tuning continues until the network is a desired size.

\subsection{Attention transfer}

\begin{figure}
    \centering
    \includegraphics[width=.5\textwidth]{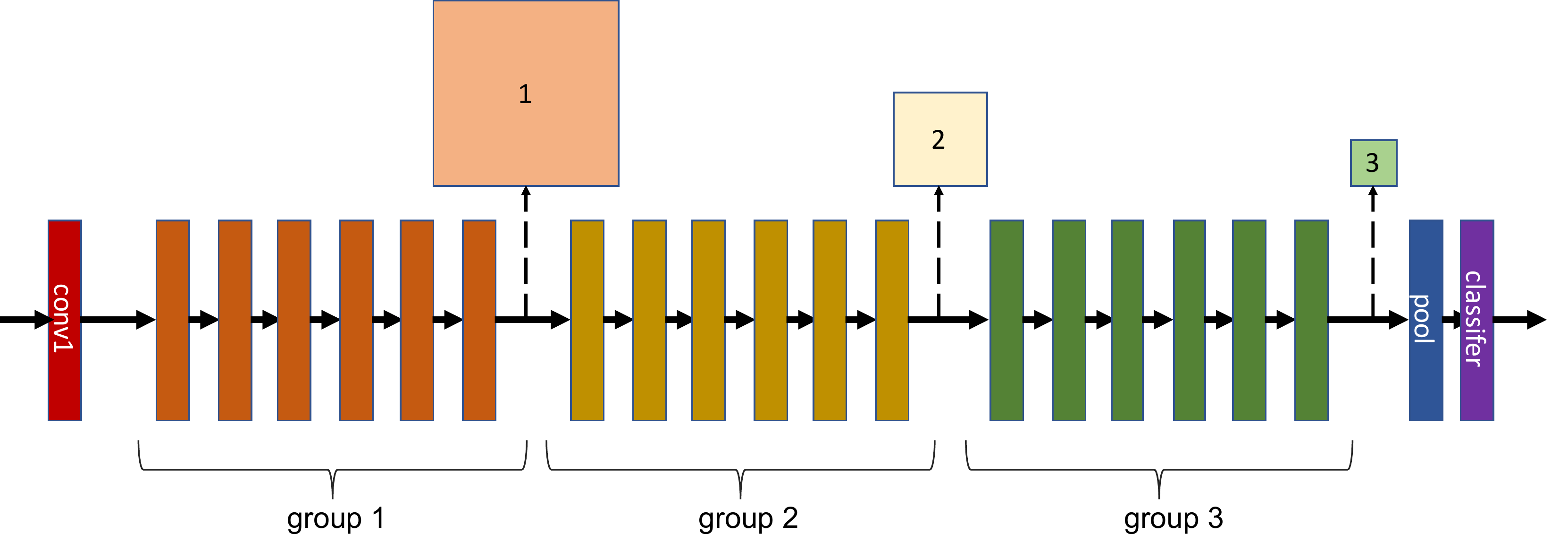}
    \caption{A block schematic diagram of a WideResNet. The network input passes through a single convolution, followed by three groups, each consisting of several residual blocks (denoted by rectangles). The output of the third group is pooled and then passes through a linear classifier. For attention transfer, attention maps are extracted at the output of each group; these are at three different spatial resolutions. }
    \label{fig:block}
\end{figure}

Attention transfer~\cite{zagoruyko2016paying} is a distillation technique whereby a student network is trained such that its {\it attention maps} at several distinct {\it attention points} are similar to those produced by a large teacher network; the intuition being that the student network is {\it paying attention} to the same things as the teacher.

Consider the WideResNet presented in Figure~\ref{fig:block}: for attention transfer, the output activations of group 1, 2, and 3 are used as attention points. To convert an activation into an attention map we square each of its elements, and then average along the channel dimension, producing a matrix where each entry corresponds to a spatial location. Finally, this map is $\ell_2$-normalised. The student network is trained with a standard cross-entropy loss and an additional term penalising the $\ell_2$ distance between each pair of teacher-student attention maps. The weight of this second term is controlled by a hyperparameter $\beta$.

A more formal definition is provided in~\cite{moonshine}: Consider a choice of layers $i =1,2,...,N_L$ in a teacher network, and the
corresponding layers in the student network. At each chosen layer $i$ of the
teacher network, collect the spatial map of the activations for channel $j$
into the vector $\aB^t_{ij}$. Let $A^t_i$ collect $\aB^t_{ij}$ for all $j$.
Likewise for the student network we correspondingly collect into $\aB^s_{ij}$
and $A_i^s$.

 Now given some choice of mapping $\fB(A_i)$ that maps each collection of the
 form $A_i$ into a vector, attention transfer involves learning the student
 network by minimising:
 \begin{equation}
 \begin{split}
 \mathcal{L}_{AT} = &\mathcal{L}_{CE}\\ & + \beta\sum_{i=1}^{N_L}  \left\lVert \frac{\fB(A^t_{i})}{||\fB(A^t_{i})||_2} - \frac{\fB(A^s_{i})}{||\fB(A^s_{i})||_2} \right\lVert_{2}, \label{eqn:atloss}
 \end{split}
 \end{equation}
 where $\beta$ is a hyperparameter, and $\mathcal{L}_{CE}$ is the standard cross-entropy loss. In~\cite{zagoruyko2016paying} the authors use $\fB(A_i)=(1/N_{A_i})\sum_{j=1}^{N_{A_i}} \aB_{ij}^2$, where $N_{A_i}$ is the number of channels at layer $i$.


\section{Discovering Performance Enhanced Student Architectures}
\label{distill}

\begin{figure*}
    \includegraphics[width=\linewidth]{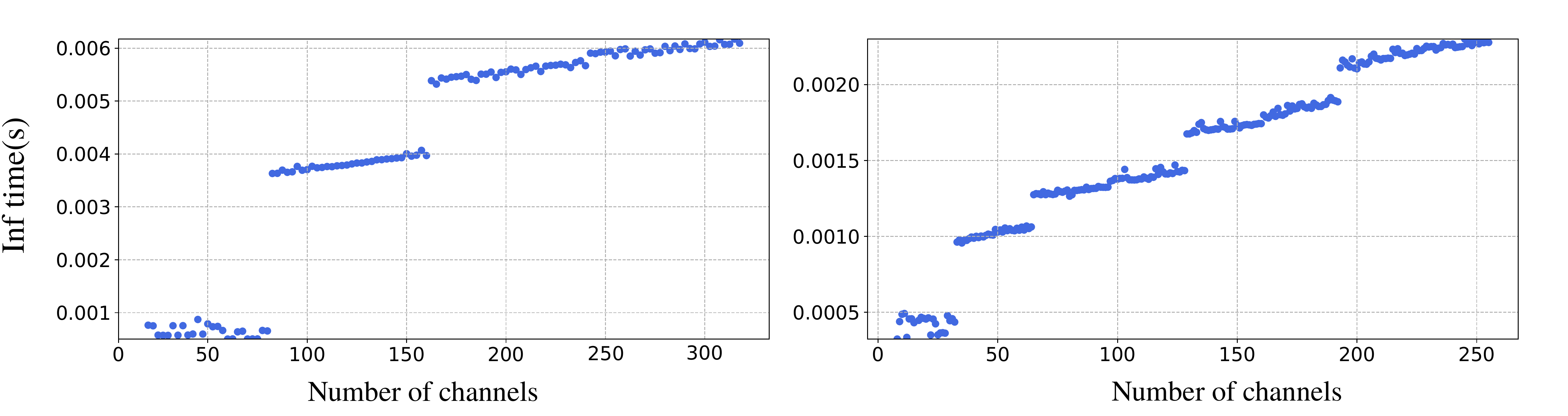}
    \caption{Further examples of the staircase pattern on different layers of the ResNet-34 architecture when deployed on the Nvidia Pascal TX2 GPU, classifying single ImageNet images. As before, we iteratively prune channels and benchmark inference time; each point on the plot therefore corresponds to a new layer initialised with a unique number of output channels.}
    \label{fig:denselayers}
\end{figure*}

We propose a two-step process for discovering performance enhanced student networks as illustrated in Figure~\ref{fig:pipeline}. We begin by developing the latency profile for each prunable layer in the network. These prunable layers are the first convolutional layers of each block, because pruning the second layer of the residual block would require us to then pad the outputs to be concatenated with the skip connection. Since we only prune the first layer of the residual blocks, each of these layers can be considered independently; that is to say, pruning one layer has no knock-on effect to later prunable layers. It is important to note that the number of input channels in the layer immediately following the pruned layer will be reduced, but the effect of this on inference time is negligible. 


\SetKwInput{kwTarget}{Target}
\SetKwInput{kwBaseModel}{BaseModel}
\SetKwInput{kwFisherModel}{FisherModel}
\begin{algorithm}[t!]
\SetAlgoLined
\kwTarget{Target hardware platform}
\kwBaseModel{Baseline pretrained model}
\kwFisherModel{Fisher-pruned BaseModel} 
student = Model()\\
 \For{i, layer in enumerate(\textbf{BaseModel})}{
  fisher\_width = FisherModel[i].layer\_width\\
  
  base\_layer = BaseModel[i]\\
  times = []\\
  \For{c in range(1 to base\_layer.layer\_width)}{
    NewLayer = Conv(base\_layer.in\_channels, c)\\
    time     = NewLayer(example\_data)\\
    times.append(time)\\
   }
   
   opt\_points = get\_outliers(times)\\
   new\_layer = Conv(base\_layer.in\_channels, nearest\_neighbour(fisher\_width, opt\_points))\\
   
   student.append(new\_layer)\\
 }
 \caption{Our student architecture discovery algorithm.
 Starting with a base model, a Fisher-pruned reduction of the base model, and a target hardware platform, we iterate over all prunable layers in the base model and construct a set of optimal points. We then adapt the pruned layer widths to their nearest optimal point, and return the resulting architecture.}
  \label{alg:pseudo}
\end{algorithm}


The latency profile of the prunable layers is given by an exhaustive empirical search of the available pruning space. For a given layer, we start with the original number of channels from the teacher model and benchmark latency for a single inference, removing a single channel at a time as illustrated in Figure~\ref{fig:denselayers}. This is an inexpensive search to perform because the layers can be considered independently, and the values of the weights of the layers do not affect inference time (meaning that there is no training involved). 

We then determine the existence of \textbf{optimal points} on the latency profile; we look for any large steps in the inference time and mark the largest channel count at each step. Intuitively, we can think of increasing the number of channels as improving the representational capacity of the network, which we hope will yield increased accuracy. Using Fisher-pruning~\cite{theis2018faster} we descend the staircase, removing channels with the lowest pruning signals. 

Once we have extracted the Fisher-pruned architecture, we adapt each layer-width choice to the nearest optimal step. This algorithm is also shown in Algorithm~\ref{alg:pseudo}. Formally, a \textbf{step} on the staircase is defined as any difference in inference time between contiguous channel counts is greater than three standard deviations from the mean difference. An optimal point on such a step is the rightmost channel count. This is labelled in Figure~\ref{fig:layer9-gpu}. In some cases---usually early layers with small feature representations---there are no discernible optimal points. In this case, we leave the layer in the state that Fisher pruning settled on. 

The existence of such optimal points is dependent on many decisions taken throughout the deployment stack; convolution algorithms, low-level primitives, matrix representations, as well as device-level parallelism characteristics (SIMD/MIMD units).

\section{CIFAR-10 Experiments}
\label{cifar}

\begin{figure*}[th!]
    \centering
    \includegraphics[width=\textwidth]{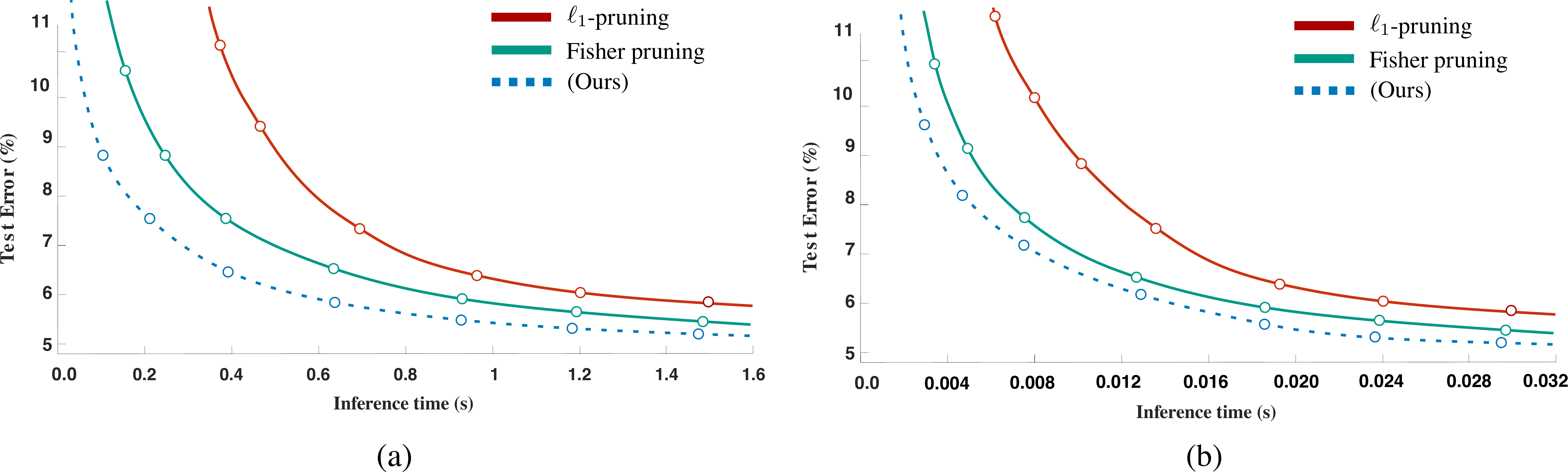}
    \caption{CIFAR-10 test error plotted against inference time for a WideResNet-40-2 on (a) the ARM Cortex CPU and (b) the Nvidia Pascal GPU. The red curve represents $\ell_{1}$ pruning, which is a common benchmark in the pruning literature. The green curve represents architectures found through Fisher pruning. Each point on the dotted curve relates to an architecture from the Fisher curve that has been adapted for the specific hardware platform and retrained via attention transfer. Our approach consistently outperforms both $\ell_{1}$-derived and Fisher-derived architectures, giving a lower error rate for a given latency budget.}
    \label{fig:wrn-40-2}
\end{figure*}

\begin{figure*}[th!]
    \centering
    \includegraphics[width=\textwidth]{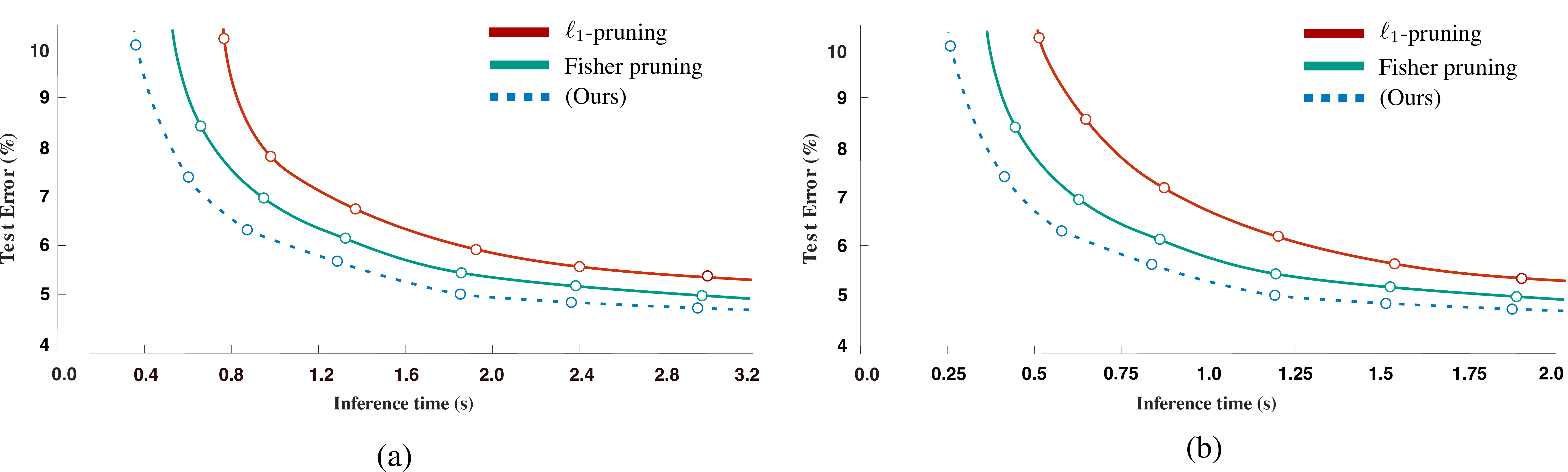}
    \caption{CIFAR-10 test error plotted against inference time for a DenseNet-100 (BC) on (a) the ARM Cortex CPU and (b) the Nvidia Pascal GPU. As in Figure~\ref{fig:wrn-40-2}, the red curve represents $\ell_{1}$ pruning, the green curve represents architectures found through Fisher pruning, and the dotted curve represents architectures derived via our method. Our approach is able to generalise from WideResNets to DenseNets, giving a lower error rate for a given latency budget than $\ell_{1}$-derived and Fisher-derived architectures.}
    \label{fig:densenet}
\end{figure*}

In this section we extensively evaluate the performance of our technique on the CIFAR-10 dataset~\cite{cifar}. We consider two exemplar networks, a WideResNet and a DenseNet, and compare our student networks to architectures found by pruning techniques at similar latency budgets. With their modular blocks and skip connections, these networks are representative of a large number of modern networks.
We show that our student networks consistently offer greater accuracy for a fixed inference time. 

A WideResNet~\cite{zagoruyko2016wide} is a modification to the traditional ResNet architecture~\cite{he2016deep} to allow for variable width. It is composed of a series of stacked blocks, each containing two $3 \times 3$ convolution layers and a skip connection. We use a WideResNet with 40 layers and a width multiplier of 2.

A bottlenecked DenseNet~\cite{huang2017densely} consists of a series of blocks each containing a $1 \times 1$ convolution, followed by a $3 \times 3$ convolution. Block outputs are concatenated and form the input to later blocks, and this encourages feature reuse to allow for powerful representations. In this work, we use a 100 layer DenseNet with these bottleneck blocks, a growth rate of 12, and a transition rate of 0.5. When performing attention transfer, attention points are placed directly after its three groups, as with the WideResNet.

\begin{table*}
    \caption{Our method compared to a pretrained ResNet-34 and a Fisher-pruned architecture at 75\% compression on ImageNet. Our method reintroduces parameters to the Fisher-pruned architecture without affecting total inference time, resulting in higher MACs/ps and reduced error.}
    \vspace{-5mm}
    \vskip 0.15in
    \centering
    \begin{tabular}{@{}l|rrrr|rr@{}}
        \multicolumn{5}{c}{} &\multicolumn{2}{c}{Nvidia TX2 (GPU)} \\
        \toprule
        Network                          & Params & MACs    & Top-1 Err.  & Top-5 Err. &   Speed   & MACs/ps  \\ 
        \hline
         Baseline ResNet-34              & 21.3M  & 4.12G   & 21.84      &  5.71        & 0.122s  & 33.77G    \\
         Fisher-pruned ResNet-34         & 5.3M   & 1.44G   & 43.43      &  18.86       & 0.038s  & 37.89G    \\
         Our ResNet-34 (with AT)         & 6.8M   & 1.58G   & 31.29      &  11.16      & 0.040s  & 39.50G    \\
        \bottomrule 
    \end{tabular}

    \label{table:speed-test}
    \vspace{2mm}
\end{table*}

Since Fisher pruning removes filters one-by-one, we can obtain a \textbf{pruning curve} comprised of a series of models at different levels of compression. Once Fisher pruning is complete we take samples from the pruning curve, evenly distributed by number of parameters. For each sample, we adapt the number of channels in each pruned layer to the nearest optimal point as defined in Section~\ref{prelim}. The network obtained from this is then randomly reinitialised and trained through attention transfer with the original network.

In order to evaluate our student networks, we compare the inference time of our samples to their Fisher-derived equivalents. As a comparison point, we also perform $\ell_1$-pruning on each original network and include similarly distributed samples from the pruning curve.

We evaluated our approach on the exemplar embedded CPU and GPU present on the Nvidia Jetson TX2 development kit. We consider the quad-core ARM Cortex-A57 CPU, and the 256-core Pascal GPU with 8GB of LPDDR4 RAM. 

The results on CIFAR-10 can be seen for WideResNet in Figure~\ref{fig:wrn-40-2} and for DenseNet in Figure~\ref{fig:densenet}. We benchmark the inference time of each sampled architecture on the ARM CPU and Nvidia GPU and plot this against the top-1 error achieved by that sample. Despite being hardware-agnostic, Fisher pruning outperforms $\ell_{1}$ pruning from a latency perspective because it is able to remove more filters for a given accuracy. Our technique, shown as the dotted curve, yields a further improvement; for each latency budget we are consistently able to offer a 0.5\% to 1\% improvement in accuracy. This holds true across both networks and both platforms, though the improvement is most notable on the CPU (due to more severe resource constraints giving larger steps in the staircase). 

\paragraph{Implementation Details} Networks are trained from scratch with SGD to minimise the cross-entropy between class labels and outputs. WideResNets are trained for 200 epochs with batch-size 128, and weight decay 0.0005. The initial learning rate is 0.1, and this is decayed by a factor of 5 every 60 epochs. DenseNets are trained for 300 epochs with batch-size 64 and weight decay 0.0001. The initial learning rate of 0.1 is decayed by a factor of 10 at epochs 150 and 225. Momentum 0.9 is used in all cases.

$\ell_{1}$-pruning, as defined in~\cite{he2017channel}, is performed in an iterative fashion. Beginning with a pretrained model,  we remove the channel with the lowest valued $\ell_{1}$-norm and then fine-tune for 100 steps. We repeat this process until every prunable channel in the network has been removed. 

We perform Fisher pruning in a similar manner. We finetune, and every 100 steps, a single channel is pruned. For WideResNet we prune the first convolutional layer in each block, whereas for DenseNet we prune the connections between the $1 \times 1 $ and $3 \times 3 $ convolution within each block. We fine-tune using the lowest learning rate the network was trained with.

 When performing attention transfer, we train the student network to minimise the loss in Equation~\ref{eqn:atloss} with $\beta$ as 1000.

\section{ImageNet Experiments}
\label{imagenet}

In this section, we apply our technique to networks trained on ImageNet~\cite{ILSVRC15}, to see whether our performance enchanced students extend to this challenging task. ImageNet is a popular computer vision dataset composed of $224 \times 224$ images, split into 1000 possible classes. The dataset has over 1 million training images, and a validation set of 50,000 examples. Test examples are unlabelled.

We evaluate the validation performance of our technique on a ResNet-34~\cite{he2016deep}. We perform Fisher pruning as above, removing a single channel every 100 steps until every prunable channel has been removed. We chose to sample a single point at a 75\% parameter compression rate and compare this to our performance enhanced student on the GPU.

The results on the validation set of ImageNet are shown in Table~\ref{table:speed-test}. With 75\% of the parameters removed, the Fisher-pruned architecture suffers a 20\% increase (to 43.43\%) in top-1 error and a 13\% increase (to 18.86\%) in top-5 error. Our approach reintroduces some parameters without affecting the inference time, with the goal of maximising the representative capacity of the network. The resulting architecture reaches 32.29\% top-1 error, a $>$10\% improvement over the Fisher-pruned architecture. 

Note that despite the large gap in accuracy, the inference time remains similar. We increase the number of parameters and MACs in the network, while maximising the throughput (MACs/ps) on our given device by leveraging the existence of optimal latency points on the channel pruning profile. Over several blocks, these width-increases accumulate into large improvements in network capacity, which translates into enhanced classification performance.

\paragraph{Implementation Details} To perform attention transfer, we train the student network to minimise the loss in Equation~\ref{eqn:atloss}. As a ResNet consists of 4 groups, we place an attention point at the end of each of these groups and set $\beta$ to 750 so the total contribution of this attentional loss is the same as for our CIFAR networks. For the teacher, we used a pretrained ImageNet model~\footnote{Pretrained models obtained from ~\url{https://github.com/pytorch/vision}}. The student network is trained using SGD and with momentum 0.9 and weight decay 0.0001 for a total of 90 epochs. The initial learning rate is 0.1, and this is decayed by a factor of 10 every 30 epochs. 

\section{Conclusion}
\label{conclusion}

In this paper we have described a simple method for discovering performance enhanced reductions of baseline, large neural networks. We have compared our technique to common pruning approaches, and demonstrated its superiority on both the CIFAR-10 and ImageNet datasets for popular networks and hardware platforms. 

It has long been understood that significant parameter redundancy exists in many deep neural networks. Now that compression techniques have matured, we are able to take advantage of insights from both developments in compiler optimisations and improvements in neural network acceleration schemes to provide an across-stack approach to optimising neural networks for specific tasks and devices. We show that taking an across-stack approach observing hardware behaviour allows us to significantly enhance the accuracy of pruned neural architectures. 

In this paper we have only considered image classification, as this is the standard task for evaluating pruning schemes. It is, however, very likely that this technique will translate to other tasks, networks, and hardware architectures. Moreover, a more thorough investigation of the staircase pattern could be helpful; future work may seek to extract the locations of optimal points based solely on hardware descriptions, avoiding the need for exhaustive empirical search. We leave such exploration for future work.

\paragraph{Acknowledgements.}
This project has received funding from the European Union's Horizon 2020
research and innovation programme under grant agreement No. 732204 (Bonseyes).
This work is supported by the Swiss State Secretariat for Education, Research
and Innovation (SERI) under contract number 16.0159. The opinions expressed and
arguments employed herein do not necessarily reflect the official views of
these funding bodies.

\FloatBarrier
\bibliographystyle{named}
\bibliography{main}

\end{document}